\documentclass[letterpaper, 10 pt, conference]{ieeeconf}
\usepackage{cite}
\usepackage{amsmath,amssymb,amsfonts}
\usepackage{algorithmic}
\usepackage{textcomp}
\usepackage{xcolor}
\usepackage[pdftex]{graphicx}
\graphicspath{{images/}}
\DeclareGraphicsExtensions{.pdf,.jpeg,.jpg,.png}
\usepackage{mathptmx} 
\usepackage{times} 
\usepackage{amsmath} 
\usepackage{amssymb}  
\usepackage{hyperref}
\usepackage{gensymb}
\usepackage{todonotes}
\usepackage[product-units=single,per-mode=symbol,range-units=single,range-phrase=\,--\,,detect-all]{siunitx}
\usepackage{balance}

\usepackage{tabularx}
\usepackage{booktabs}
\usepackage{silence}
\usepackage{caption}
\usepackage{subcaption}

\newcommand{\reffig}[1]{Fig.~\ref{#1}}
\newcommand{\reftab}[1]{Tab.~\ref{#1}}

\newcommand{\refsec}[1]{Sec.~\ref{#1}}

\def\BibTeX{{\rm B\kern-.05em{\sc i\kern-.025em b}\kern-.08em
    T\kern-.1667em\lower.7ex\hbox{E}\kern-.125emX}}

\title{\LARGE \bf Redefining Recon: Bridging Gaps with UAVs, 360° Cameras, and Neural Radiance Fields}

\author{Hartmut Surmann, Niklas Digakis, Jan-Nicklas Kremer, Julien Meine, Max Schulte, Niklas Voigt}
\thanks{Corresp. author: {\tt\small hartmut.surmann@w-hs.de}}
\thanks{Corresponding author's: {\tt\small hartmut.surmann@w-hs.de}}
\thanks{This work is funded by the German Ministry of Education and Research (BMBF), grant No. 13N16478, project “E-DRZ: Establishment of the German Rescue Robotics Center” (DRZ) \url{https://www.rettungsrobotik.de}}

\begin{document}

\maketitle

\begin{abstract}
In the realm of digital situational awareness during disaster situations, accurate digital representations, like 3D models, play an indispensable role. To ensure the safety of rescue teams, robotic platforms are often deployed to generate these models. In this paper, we introduce an innovative approach that synergizes the capabilities of compact Unmaned Arial Vehicles (UAVs), smaller than 30 cm, equipped with 360° cameras and the advances of Neural Radiance Fields (NeRFs). A NeRF, a specialized neural network, can deduce a 3D representation of any scene using 2D images and then synthesize it from various angles upon request. This method is especially tailored for urban environments which have experienced significant destruction, where the structural integrity of buildings is compromised to the point of barring entry—commonly observed post-earthquakes and after severe fires. We have tested our approach through recent post-fire scenario, underlining the efficacy of NeRFs even in challenging outdoor environments characterized by water, snow, varying light conditions, and reflective surfaces.
\end{abstract}

{\bf keywords}: Small UAVs, 360°-Panorama, visual monocular SLAM, UAV, Rescue Robotics, NeRF

\section{Introduction}
\label{sec:introduction}

Robots are extremely useful to the operations of emergency forces, providing invaluable support across air, water, and land\cite{MurphyTK16}. While unmanned aerial vehicles (UAVs) have found their niche in outdoor reconnaissance, ground robots face a multitude of challenges. Though equipped with advanced cameras and laser scanners capable of producing extensive 3D models, their efficacy plummets in cluttered terrains. Even with just a moderate amount of debris, these robots are severely hindered, especially within urban and indoor settings. Rescue missions frequently present environments obstructed with rubble and debris, conditions that these ground robots find nearly insurmountable. UAVs become the most viable solution for assessing these situations\cite{Pratt2008}. Our project proposes the use of compact, lightweight UAVs ($<$ 30 cm) that have been available for the past three years. These digital FPV (First-Person View) UAVs, equipped with radio support, high-quality cameras, and excellent flight characteristics, provide rapid imagery of unstable environments. They can have the added advantage of being user-friendly, not just for UAV specialists but also for minimally trained emergency responders \cite{9597869}.


\begin{figure}[ht]
\begin{center}
\begin{minipage}{0.5\columnwidth}
  \includegraphics[height=3.25cm]{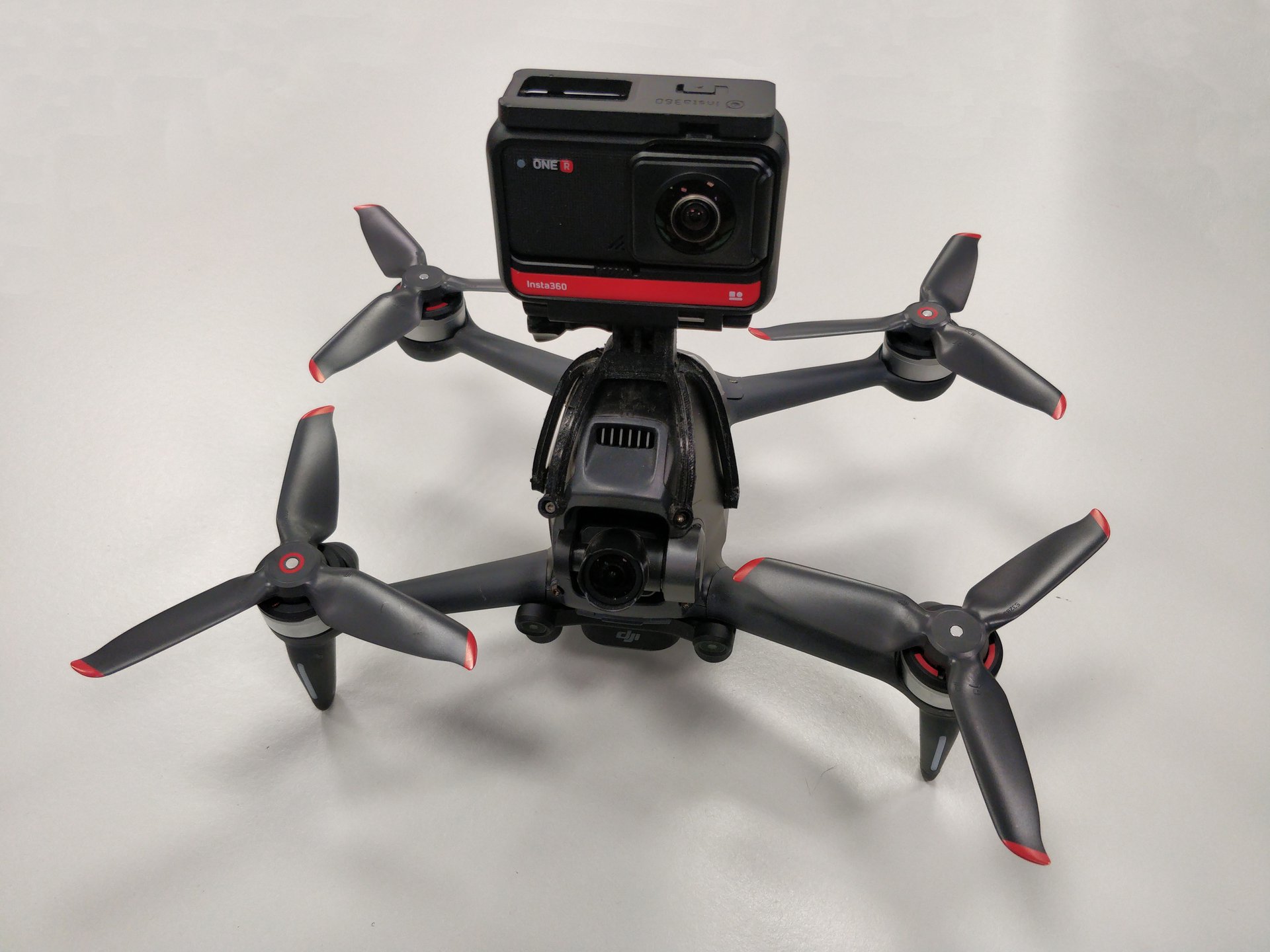}
\end{minipage}%
\begin{minipage}{0.5\columnwidth}
  \includegraphics[height=3.25cm]{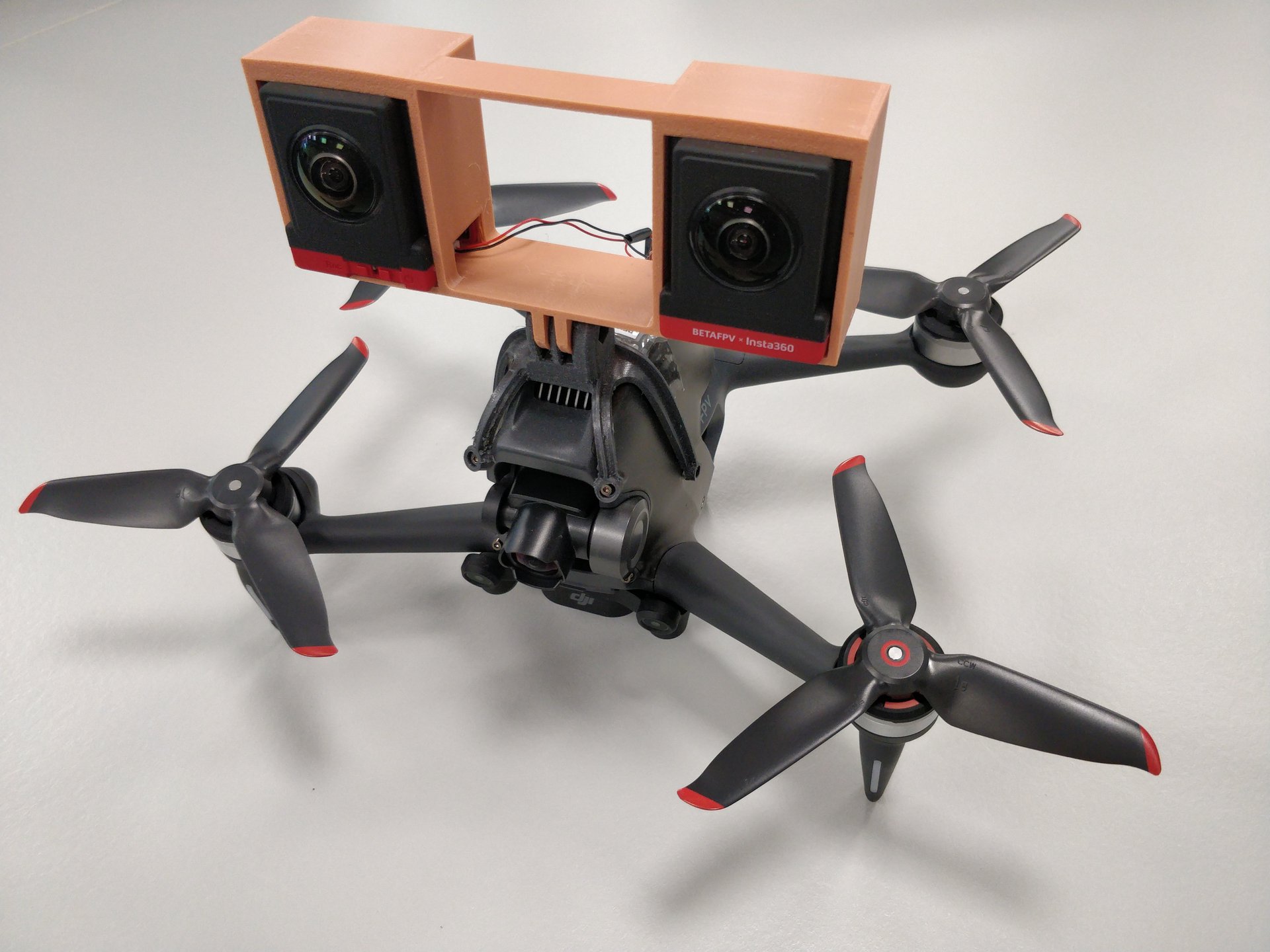}
\end{minipage}
\begin{minipage}{0.5\columnwidth}
  \includegraphics[height=3.25cm]{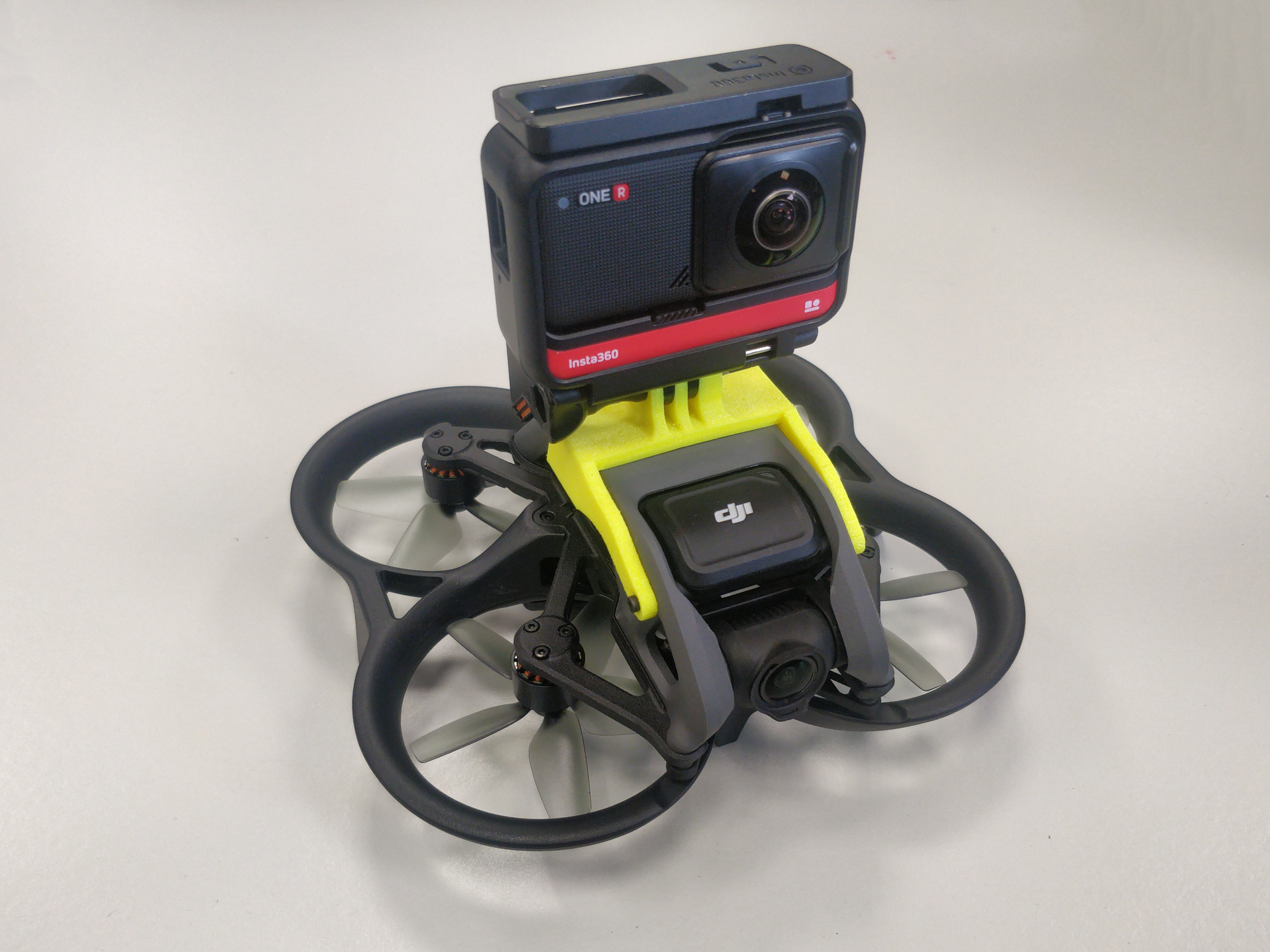}
\end{minipage}%
\begin{minipage}{0.5\columnwidth}
  \includegraphics[height=3.25cm]{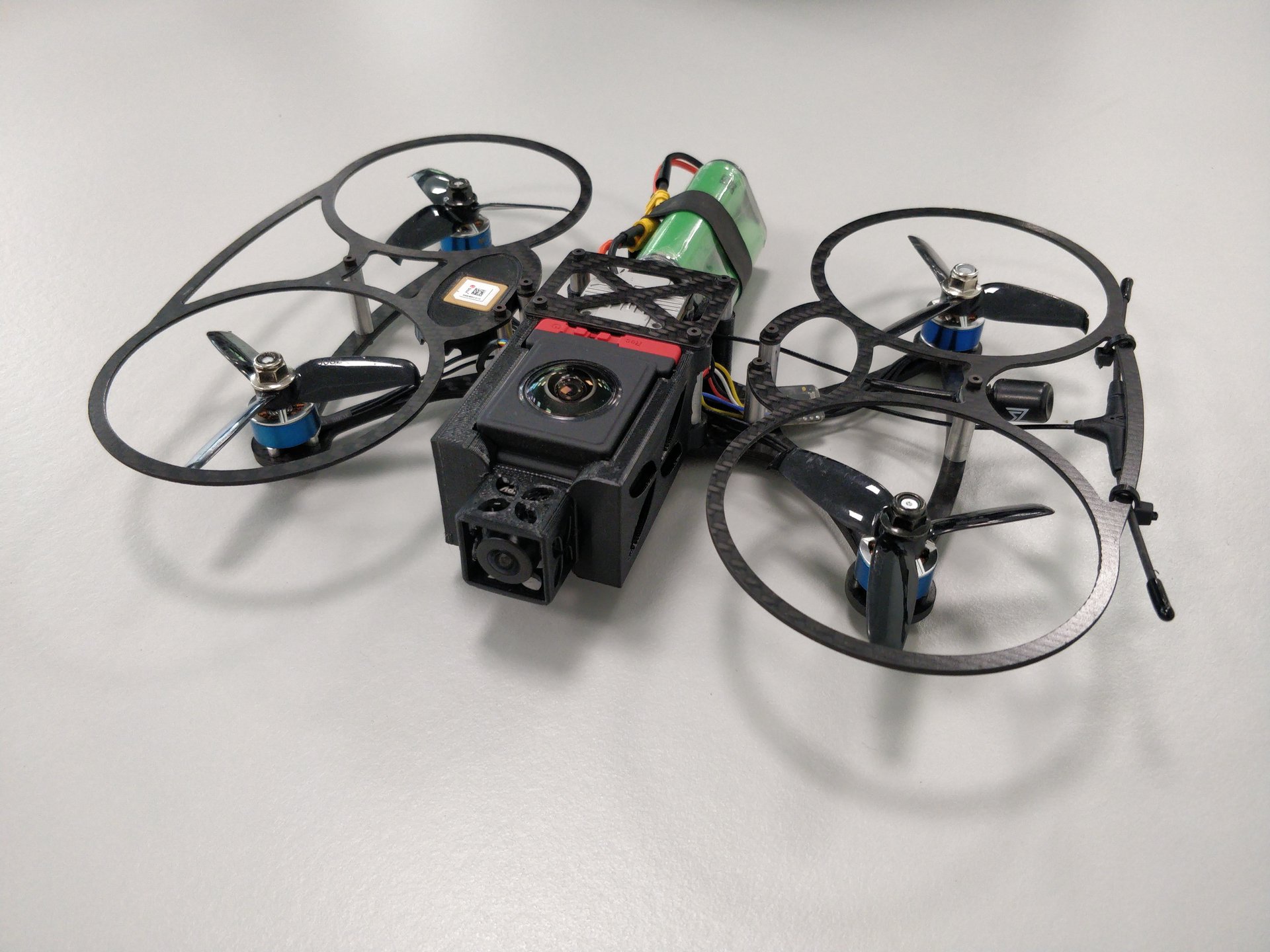}
\end{minipage}
\end{center}
  \caption{Examples of small UAVs. Top: DJI FPV adapted with one and two 360° cameras. Bottom: DJI Avata and a self-constructed invisible UAV, both with 360° cameras.}
 \label{fig:small-uavs}
\end{figure}

UAVs, as of recent times, have been ubiquitously adopted by many fire departments across Germany\cite{Feuerwehrverband2023}. According to our observations, they are usually used for reconnaissance, mainly only analyzing the real-time data from the UAV's cameras. This indicates that there is often a lack of systematic evaluation and deeper processing of this vast data. To bridge this gap, our project introduced ARGUS\footnote{\url{https://github.com/RoblabWh/argus}}, a tool designed to streamline data analysis. Within this tool, we've improved data archiving, integrated available thermal imaging, generating orthophotos, and incorporated artificial intelligence for targeted objects and hazard identification.  As this tool continues to evolve in response to the needs of fire and rescue services, more and more processes, and algorithms for processing the captured UAV Data got tested and evaluated in real-world missions. This paper dives deeper, exploring enhancements in digital representations and 3D modeling. 

Large UAVs, equipped with multiple sensors especially tailored for rescue forces, such as the DJI M30T and M300 models, have quickly established themselves in Germany over the last years \cite{Drohneneinheit2023}. Yet, their deployment is largely confined to open outdoor spaces. In environments like industrial units or damaged buildings, these UAVs struggle with their bulky size, extensive obstacle avoidance features, and lack of GPS support indoors. The vulnerability of these UAVs in such settings underscores the need for alternative solutions that can withstand the rigorous demands of indoor operations. This can be easily met in the form of versatile small UAVs from the FPV scene, equipped with additional sensors and cameras to suit them best for use in rescue missions.

Recognizing the need for comprehensive visual data, we advocate for equipping these UAVs with 360° cameras. This ensures an uninterrupted view of the surrounding environment, significantly enhancing the UAV's localization quality, which is crucial for visual 
simultaneous localization and mapping (SLAM).
During disaster operations, UAVs navigate challenging terrains, from dark spaces to bright open areas, often entering through narrow openings like broken doors or windows. Conventional cameras with a field of view of under 100° often disrupt visual localization, making remodeling a scene unattainable\cite{MurphyTadokoro2019}. Our proposed integration effectively addresses this issue, ensuring consistent visual data collection even in such demanding scenarios.

By embedding 360° cameras within the UAV framework, we capture unhindered imagery of incident sites. This provides a favorable foundation for post-processing, particularly for SLAM algorithms, as it removes the peril of misplacing tracking markers in constricted spaces. The compact UAVs we discuss in this paper (see \reffig{fig:small-uavs}) present an affordable and repairable option for high-risk operations. We further highlight the advantages of these UAVs, emphasizing their unmatched capability to capture obstruction-free 360° panoramic shots.

Finally, beyond the introduction of enhanced UAV technologies, our work integrates Neural Radiance Fields (NeRFs) for data processing. NeRFs stand out, quickly translating 360° imagery into preliminary scene representations. They often outpace conventional methods like point cloud and mesh model generation in both speed and perceived accuracy. Throughout this paper, we present our employed NeRF generation methods and conduct a comparison with the traditional approaches mentioned before. Our findings highlight the potential of NeRFs as a valuable tool for critical rescue scenarios.

The outline of this paper is as follows. In the next section, we describe the state of the art and relevant preliminary work to this paper. In \refsec{sec:deployment} we describe the system approach consisting of the 360° UAVs, the localization of the 360° images, and the representation using the NeRF models. The fourth section (\refsec{sec:results}) then evaluates the system approach using two examples. The first one is from a deployment at real fire in Essen 2/2022 \cite{surmann2022lessons} and the second is a hard-to-process data set with snow and reflective surfaces from 2/2023 at the Westphalian University of Applied Sciences (WHS). The videos for these are available on YouTube\footnote{Essen: \url{https://www.youtube.com/watch?v=Pd2__gm0nUE},\\  WHS: \url{https://www.youtube.com/watch?v=zH0qzGDpJW8}}

\section{Related Work}
\label{sec:related-work}

The integration of 360° cameras with UAVs has gathered some interest in the research community. Numerous papers have detailed the applications of this technology in UAVs and have pinpointed existing challenges as well as opportunities for further enhancement.

For instance, Previtali et al.\cite{Previtali2023} present an innovative methodology that utilizes both ground and aerial data to create comprehensive 3D models of small-town historic urban environments. Santano et al.\cite{8346262} analyzed how UAV movements correlate with visual data captured using a retrofitted Samsung Gear 360 camera. The study included a thorough evaluation of the 360° footage quality, such as bitrate, frame rate, dynamic range tests, file compression, and stitching software. Despite the valuable contributions of these studies, the challenge of the UAV being visible in the imagery persist. This is a considerable limitation as it can interfere with subsequent data processing and scene interpretation. 

Caroti et al.\cite{isprs-archives-XLII-3-W4-137-2018} focused on using caged micro-UAVs with 360° cameras for disaster management in challenging environments. The paper acknowledges challenges such as changing image framing due to the cage wires on the field of view, and tests various photogrammetric processing procedures to evaluate their efficacy in creating accurate 3D models without visibility of the cage and the UAV.

Attempting to mitigate the problem of visibility of the UAV, a research paper by Kim et al.\cite{kim2019UAV360degree} introduce a novel UAV design model and video merging technique that facilitates 360° real-time video recording while making the UAV virtually invisible. In a similar manner Zia et al.\cite{8661954} presented an automated system to create high-quality, full 360° panoramas from images taken by fisheye cameras mounted on a UAV. However, this solution required multiple cameras recording from various angles, complicating the system's design, operation and cost.

In contrast, our work stands out by employing a single 360° camera mounted in such a way that the UAV remains invisible in the acquired imagery. This approach simplifies the system's design and operation while ensuring seamless, comprehensive coverage of the environment. Moreover, where previous work primarily focused on outdoor applications of UAVs our research further extends this technology's utility by addressing the challenges specific to indoor and GPS-denied environments.

Despite advancements in the integration of 360° cameras with UAVs, improving the quality and interpretability of the acquired imagery remains a challenge. Advanced image processing techniques play a crucial role in converting this raw footage into meaningful reconstructions of the environment. Enter Neural Radiant Fields (NeRFs), a breakthrough in the field of Neural View Synthesis\cite{mildenhall2020nerf}. Through the utilization of Deep Neural Networks, NeRFs learn a 3D representation of a specific environment, generating highly accurate arbitrary views. To accomplish this, the model has to be trained on a localized sparse set of images, capturing the environment.
Similar to the plenoptic function introduced by Adelson und Berger in 1991 \cite{AdelsonBerger1991}, a NeRF approximates the density and radiance of any point in a 3D space based on a 5D input vector consisting of the spatial location ($x$,$y$,$z$) and the viewing direction ($\theta$, $\phi$) \cite{mildenhall2020nerf}. To render a scene, a ray is sent through every pixel of the viewing plane, along which the neural network is queried multiple times. The values along each ray get summed up based on their density to form the color of the specific pixel \cite{mildenhall2020nerf}.
Another breakthrough was instant-ngp\cite{mueller2022instant}, which, by incorporating an input hash encoding, reduced the computation time significantly by several orders of magnitude \cite{mueller2022instant}.
Mega-NeRF, as presented by Turki et al.\cite{Turki_2022_CVPR}, represents a noteworthy advancement in the UAV-centric NeRF domain. While tailored for large-scale 3D rendering, the authors suggest its potential for UAV-driven search-and-rescue applications, although this hasn't been explicitly tested. Given the vast terrains typical of search-and-rescue missions, Mega-NeRF's ability to handle large datasets could be a valuable asset for UAV surveillance in such scenarios.

In conclusion, while the integration of 360° cameras with UAVs is a growing research area, it is clear that there is a need for a simple yet effective solution to the issue of the UAV's visibility in the captured images. Our solution uses a single 360° camera and ensures the UAV's invisibility, which helps with data processing. Furthermore, it's suitable for both outdoor and indoor environments without GPS. By integrating NeRFs, we've actualized the use of these invisible UAVs in real-world rescue scenarios. By rendering precise 3D representations from the NeRF technology, our system provides detailed and accurate aerial perspectives, enhancing the precision and efficiency of rescue missions.

\section{Deployment}
\label{sec:deployment}


\subsection{Hardware}
\label{sec:hardware}


\subsubsection{UAVs}
\label{sec:uavs}

Deployed UAVs may vary depending on the scenario. Operations can be broadly categorized into indoor and outdoor scenarios. Outdoor areas are surveyed using UAVs like the DJI Matrice M30T or the DJI Mavic. Besides their extended flight time, these UAVs are advantageous due to their resistance to wind and weather conditions. However, there is room for improvement when it comes to indoor operations. An operation in Berlin \cite{9597677} demonstrated that without additional equipment, UAVs couldn't enter a building through its roof windows. The Matrice/Mavic UAVs are too large for indoor environments. Furthermore, a propeller guard is essential for indoor use to prevent general damage should the UAV collide with objects. For navigating through small openings or windows, smaller UAVs like the DJI Avata or self-constructed FPV UAVs, which are just a quarter of the size of a Mavic, are more appropriate. In areas without GPS, such as indoor environments, UAVs face challenges in maintaining stability and position. This is due to the lack of GPS signal, electromagnetic interference, unpredictable air currents, and the diverse materials and confined spaces that challenge obstacle detection. To manage these challenges, UAV stabilization is either handled through optical sensors or manually by the pilot.

As professional and off-the-shelf UAVs could not satisfy all of our demands for a UAV in an indoor setting, we investigated more into the field of FPV racing UAVs. Since a lot of these UAVs can be built, repaired, and altered with a wide variety of modular parts, coming from a do-it-yourself scene, we were able to develop and test multiple UAVs, each with protected propellors, enclosed inside of the UAVs frame, while staying below a size of 30 cm. Utilizing the DJI Digital HD FPV System system with a latency of less than 30 ms, we can deliver a precise real-time perspective of the scenario for both the pilot and the ground station.

Within our UAV system architecture, the flight controller incorporates the ArduPilot\footnote{\url{https://ardupilot.org}} software, an advanced open-source autopilot system renowned for its versatility, particularly in the domains of autonomy and detailed mission planning. This platform facilitates the integration of a diverse array of sensors and actuators, enabling the programming for autonomous operations. Given specific operational requirements and intentions, this software is adept at navigating the complexities of indoor flight dynamics. Since we adapted UAVs from FPV Racing, they had to mostly be controlled completely manually by the pilot. To mitigate pilot workload and enhance flight support, we explored a range of sensors, including LiDAR and tracking cameras, to leverage ArduPilot's assistive capabilities. Extensive testing occurred at the German Rescue Robotics Center (DRZ), simulating real-world conditions. Notably, the integration of the Intel RealSense T265 yielded the best outcomes, ensuring UAV stability even in challenging lighting conditions and foggy environments. However, limitations arose from its size and placement, protruding the view of the installed 360° camera. Furthermore, since the Intel RealSense T625 was discontinued from the manufacturer, without a suitable successor, this configuration was not employed in subsequent real operations. In the absence of an appropriate alternative, our self-constructed UAVs were maneuvered manually by our skilled pilots during actual missions.

A comparison of the small UAVs we used can be seen in \reftab{tab:uavs}.

\begin{table}[ht]
\centering
\caption{Deployed small UAVs}
\renewcommand{\tabularxcolumn}[1]{>{\centering\arraybackslash}m{#1}}
\begin{tabularx}{\columnwidth}{|X|XXXX|}
\hline
                & A             & B             & C                 & D                 \\
\hline
Size            & 180x180mm     & 320x380mm     & 420x450mm         & 200x270mm         \\
\hline
Weight          & 570g        & 955g          & 894g              & 425g              \\
\hline
Flight time      & $\sim$9min    & $\sim$15min   & $\sim$10min       & $\sim$13min       \\
\hline
Flight assistance (indoor)      & yes   & yes   & no       & possible        \\
\hline
Propguard           & incl.         & excl.         & no                & incl.             \\
\hline
Occlusion      & 33\% bottom   & 33\% bottom   & 0\%              & 0\%               \\
\hline
\end{tabularx}
\label{tab:uavs}
\\
(A) DJI Avata with Insta360 One R               
(B) DJI FPV with Insta360 One R                 
(C) DJI Mavic 1 Pro with Insta360 Sphere        
(D) Self-constructed invisible UAV with BetaFPV SMO 360    
\end{table}









\subsubsection{Image Sensors}
\label{sec:image-sensors}

For quick aerial surveys in outdoor areas, we mostly utilize the integrated cameras of the DJI Mavic or Matrice UAVs. The high-resolution images captured contain embedded GPS data, which can be directly used by GIS systems like ODM or our own tool ARGUS.

As mentioned in the introduction, 360° cameras offer many benefits, particularly for visual slam algorithms. Additionally, they allow the pilot to use a forward-facing first-person camera exclusively for maneuvering the UAV so that it does not have to be directed in a specific direction to investigate regions of interest (ROIs) during the flight as the viewing direction can be individually chosen later while observing the recorded 360° video.  
Throughout the iterations of our self-constructed UAVs, we've experimented with different cameras. We started with an Insta 360 One R (weight: 165 g). However, to reduce weight and maximize flight time, we transitioned to the BetaFPV SMO 360 (weight: 55 g).


\begin{figure}[ht]
  \centering
  \includegraphics[width=\columnwidth]{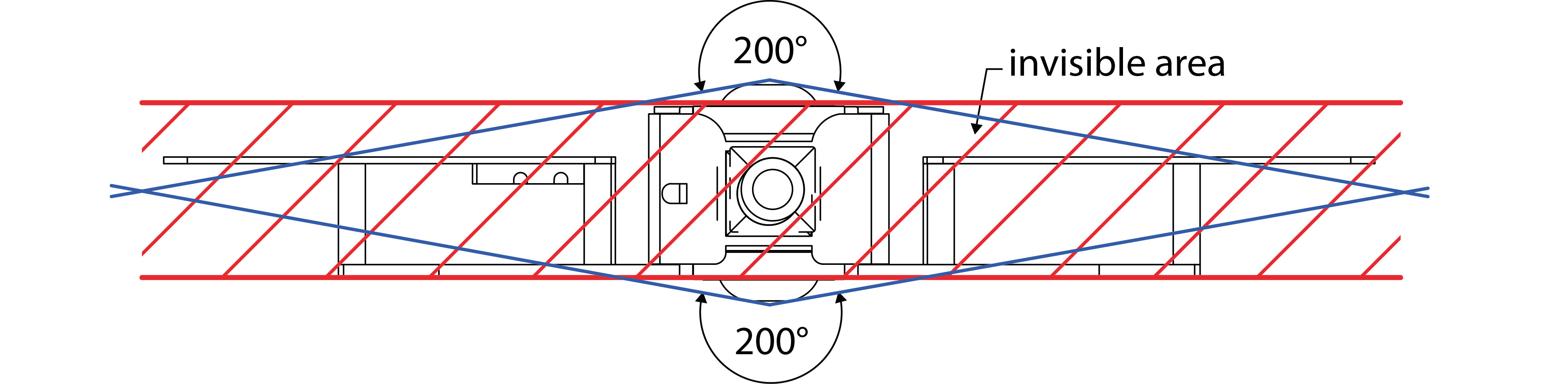}
  \caption{Schematic drawing illustrating the functioning of the self-constructed invisible uav.}
  \label{fig:invis-drone}
\end{figure}

\begin{figure}[ht]
  \begin{subfigure}[b]{0.495\columnwidth}
    \includegraphics[width=\columnwidth]{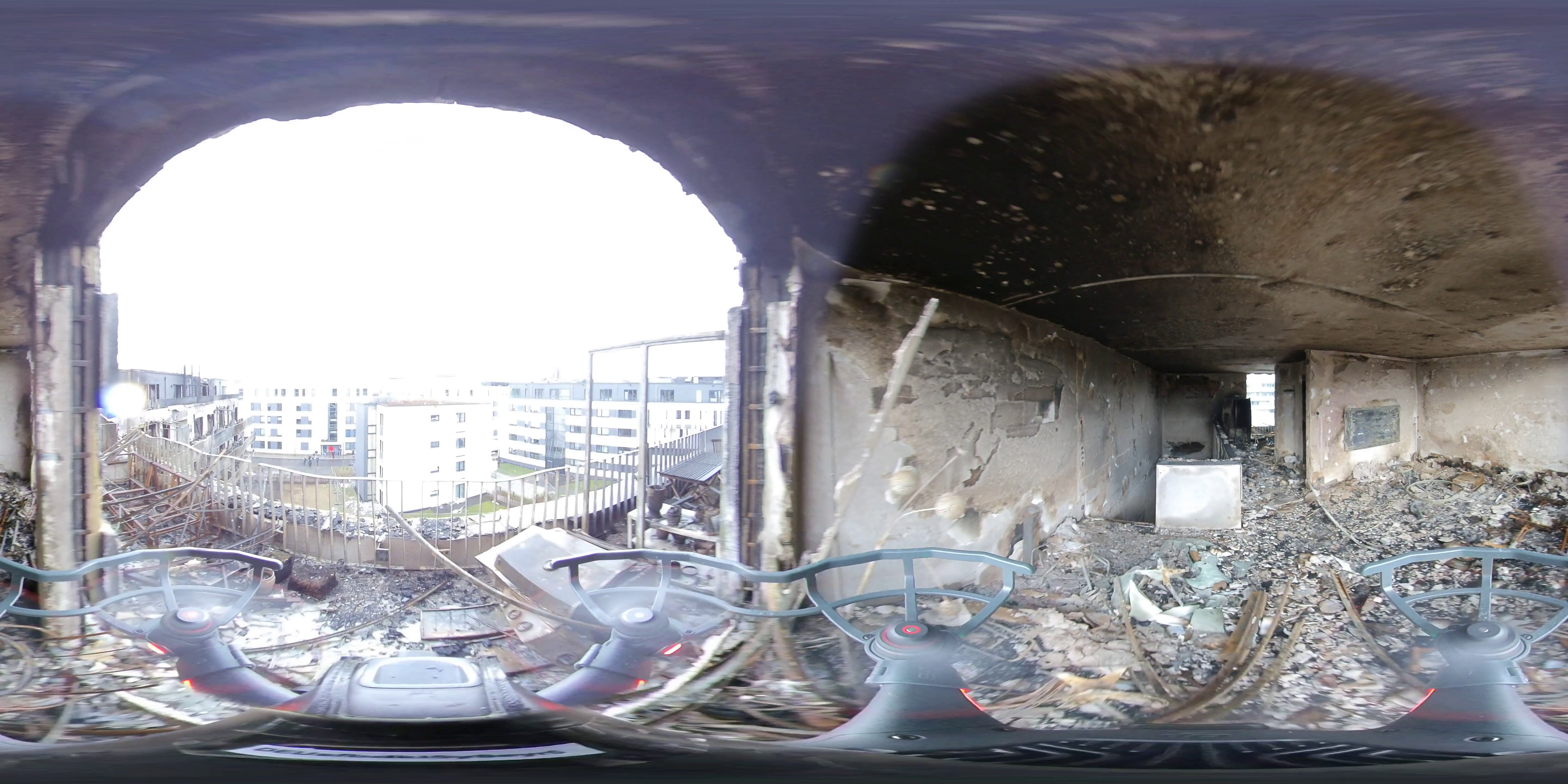}
    \caption{Burned-out building in Essen with the camera mounted on top of a DJI FPV.}
    \label{fig:pano-essen}
  \end{subfigure}
  \hfill
  \begin{subfigure}[b]{0.495\columnwidth}
    \includegraphics[width=\columnwidth]{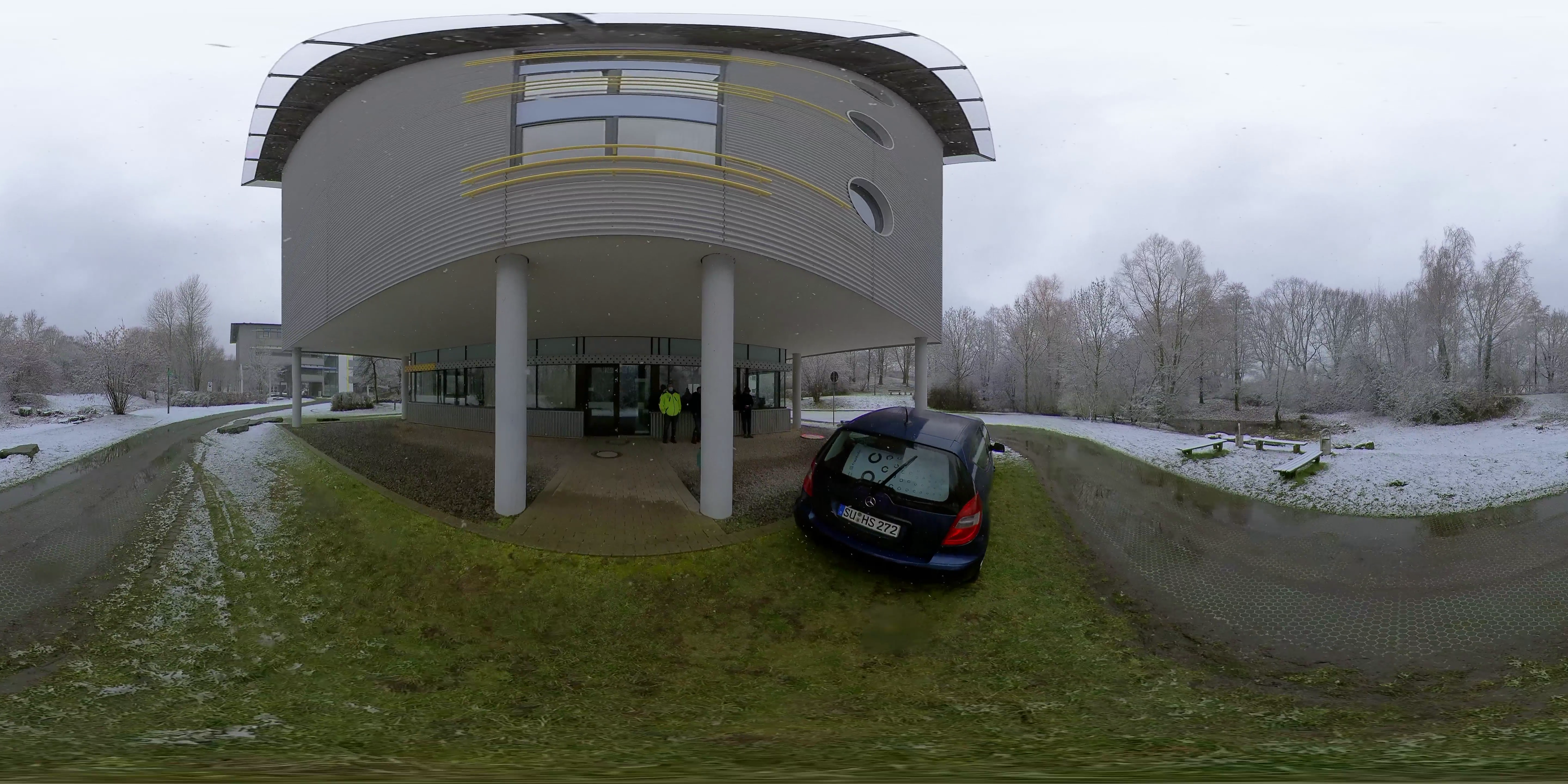}
    \caption{Westphalian University of Applied Sciences flight of self-constructed invisible uav.}
    \label{fig:pano-diyd}
  \end{subfigure}
  \caption{Panoramas of two UAVs with 360° cameras.}
  \label{fig:pano-diff}
\end{figure}

The utilization of 360° panoramic cameras is incredibly valuable as they allow us to capture the entire surrounding environment at any point in time. Therefore minimizing any obstructions by the UAV itself is crucial, which is why the 360° camera got mounted inside the frame of our self-constructed UAV, with the lenses clearing the edges of its frame (as seen in \reffig{fig:invis-drone}), leading to an unoccluded 360° image (as seen in \reffig{fig:pano-diyd}). The used commercial UAVs got equipped with 360° cameras by specifically constructed 3D printed mounts, mostly placing the camera on top of the UAV, leading to  panoramas partially occluded by the UAV (see \reffig{fig:pano-essen}). This shows the benefit of the images taken from our self-constructed UAV including more information about the ground, which is especially important in indoor scenarios.

\subsubsection{Ground Station}
\label{sec:ground-station}

In operational scenarios, providing an appropriate interface to both the pilot and the operations command is crucial for flight and route planning. This interface aids in delivering data, telemetry, and ultimately, information crucial for situation assessment. 
The DJI UAVs are controlled with their proprietary controller, either directly or by previously planning a flight mission. Through the DJI controller, we can share live feed with the operating forces via an RTMP stream. The self-constructed UAVs can additionally be monitored and controlled through software like MissionPlanner or QGroundControl using a MavLink connection.
For processing the acquired data in the field, a laptop is used equipped with an Intel Core i7-10750H, an GeForce RTX 2080 Super 8GB, and 32 GB of RAM. This laptop presents a realistic mobile setting that could also be used by fire departments and will be used for later testing in this paper.

\subsection{Procedure}
\label{sec:procedure}

\begin{figure}
    \centering
    \includegraphics[width=\columnwidth]{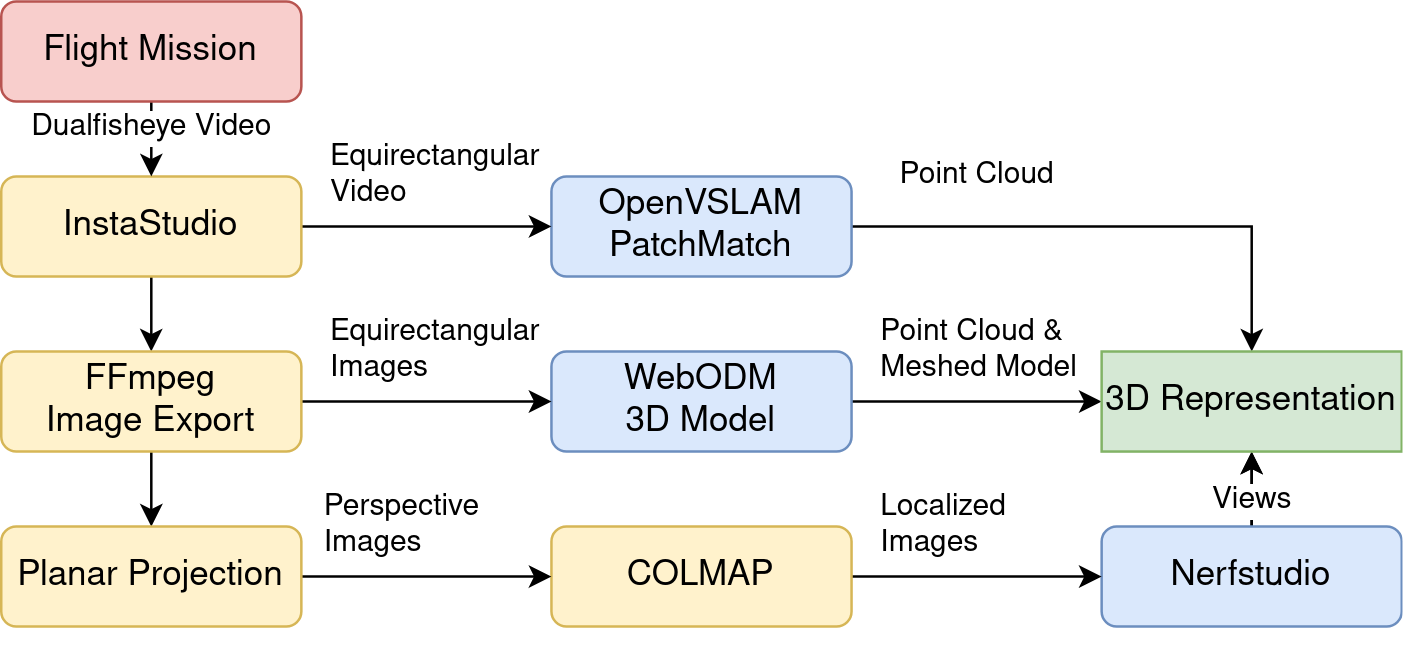}
    \caption{Diagram depicting data processing using different methods after the flight mission. Preprocessing steps are highlighted in yellow, main processing in blue, and the flight mission and results are denoted in red and green, respectively.}
    \label{fig:data-proc}
\end{figure}

Depending on the situation, initial reconnaissance usually includes mapping flights and additional flights over local ROIs, often using our self-constructed 360° UAVs.
After capturing the relevant environments with our 360° UAVs, the videos are available in the dual fisheye format. To further process them with different software, they need to be converted into the equirectangular format.
To achieve this, we stitch the videos using the Insta360 Studio software provided by the manufacturer. Currently, this process still requires manual intervention. The start and end of the flight need to be defined before initiating the stitching process. For orthophoto creation, we use WebODM or our own mapping approach implemented within ARGUS.
We employ WebODM and NeRFs to create our 3D representations (\reffig{fig:data-proc}), achieving both accuracy and detail. This synergy serves various use cases, yielding superior models tailored to operational requirements.

\subsection{Utilized Methods}
\label{sec:utilized-methods}






\begin{figure*}
    \centering
    \begin{subfigure}[b]{0.329\textwidth}
        \includegraphics[width=\textwidth]{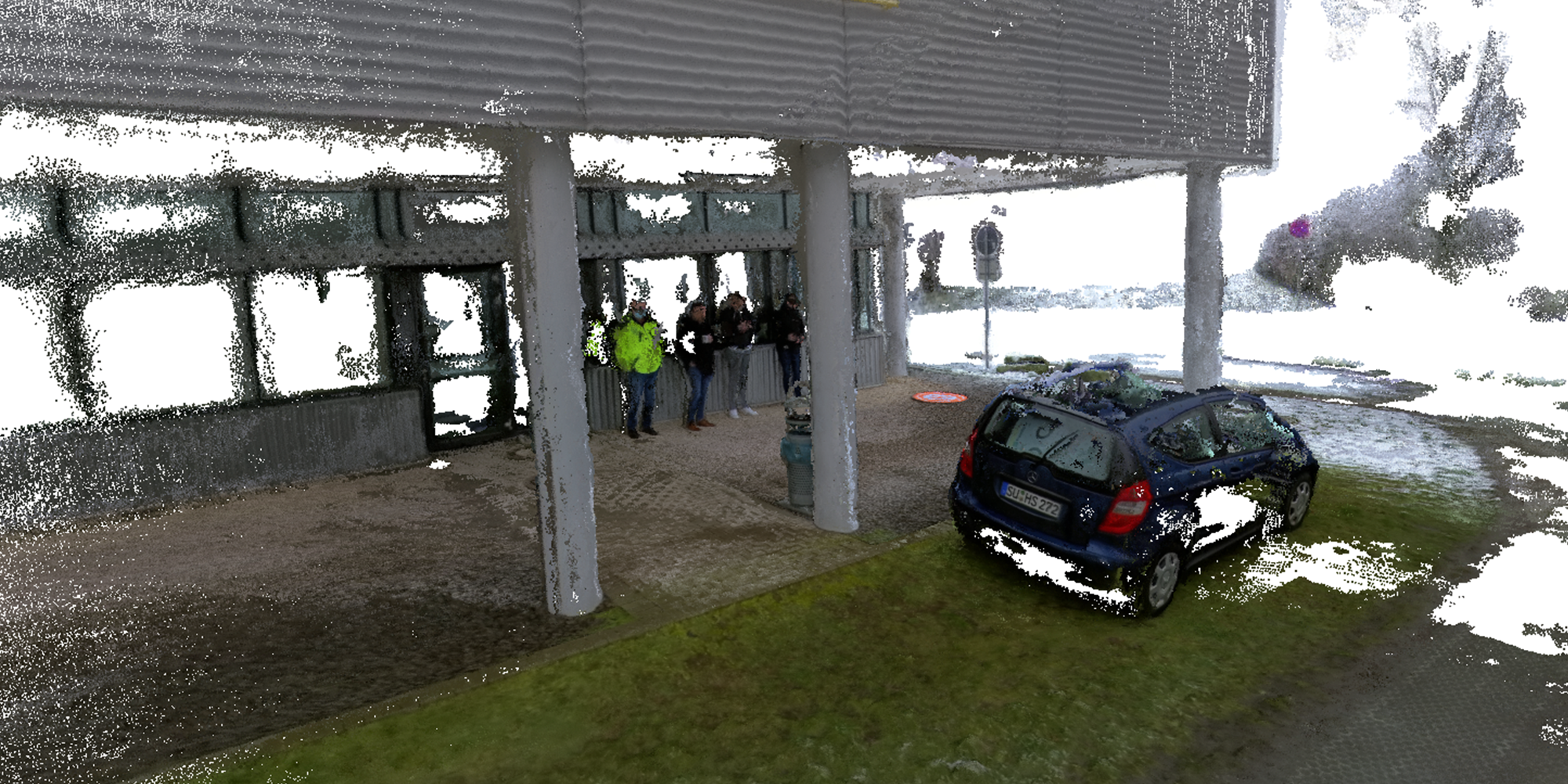}
        \caption{WebODM Point Cloud}
        \label{fig:odm-diyd}
    \end{subfigure}
    \hfill
    \begin{subfigure}[b]{0.329\textwidth}
        \includegraphics[width=\textwidth]{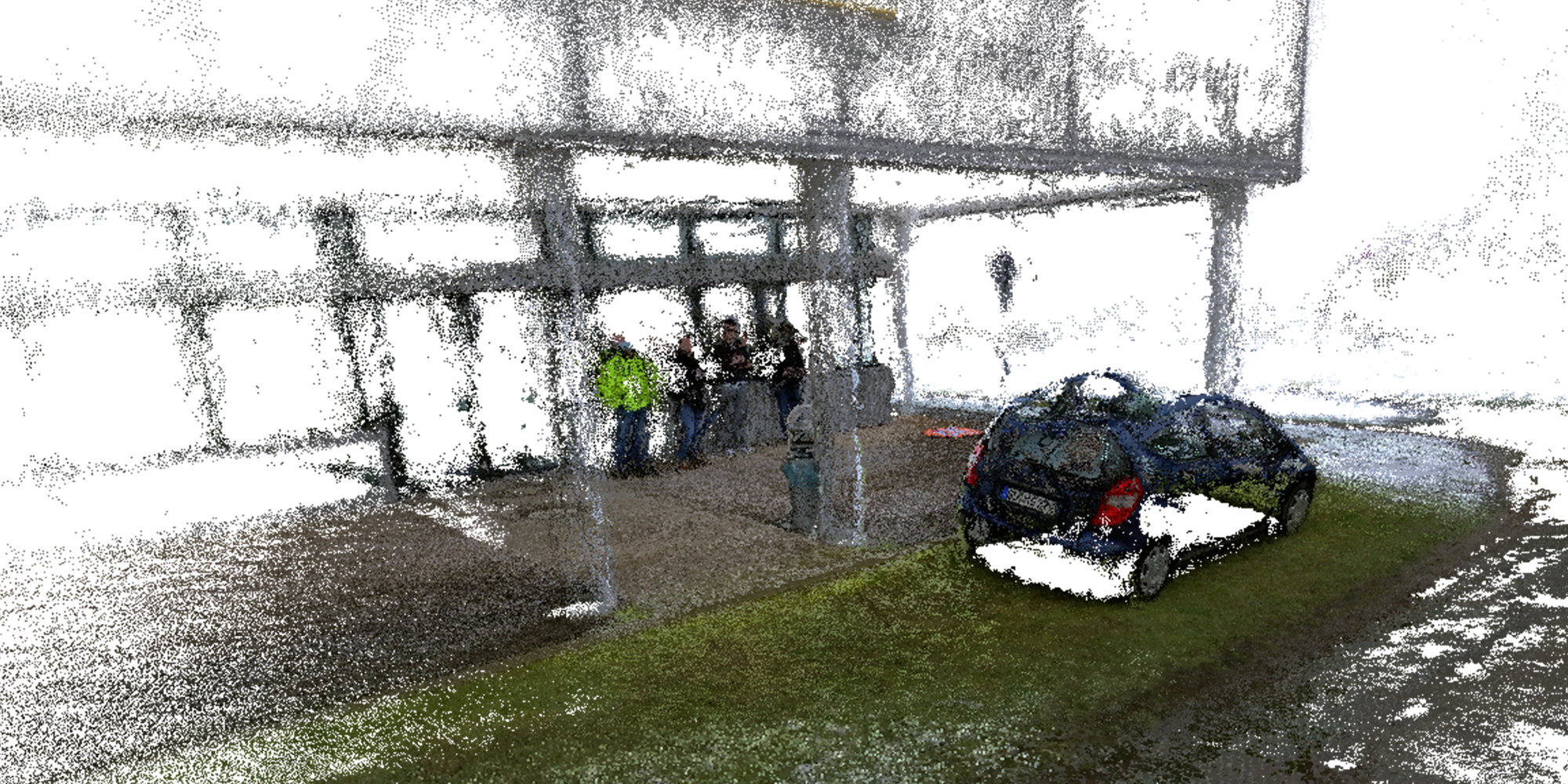}
        \caption{PatchMatch-Stereo-Panorama}
        \label{fig:pmd-diyd}
    \end{subfigure}
    \hfill
    \begin{subfigure}[b]{0.329\textwidth}
        \includegraphics[width=\textwidth]{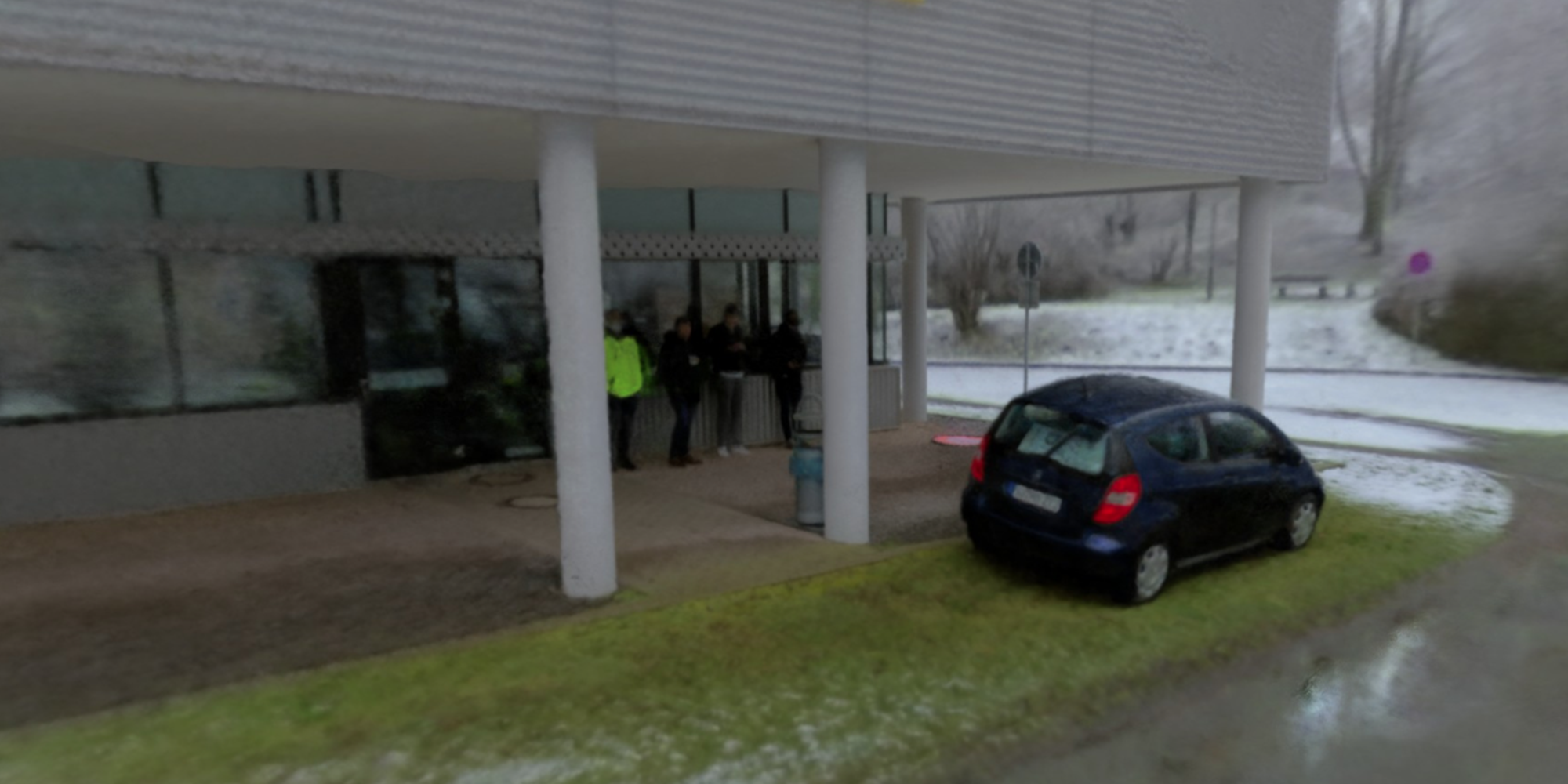}
        \caption{Nerfstudio Nerfacto}
        \label{fig:nerf-diyd}
    \end{subfigure}
    \caption{Assessment of the self-constructed invisible UAV using different methods at the Westphalian University of Applied Sciences.}
    \label{fig:diyd-results}
\end{figure*}

\subsubsection{WebODM}
\label{sec:webodm}
Our initial approach involved utilizing WebODM\footnote{\url{https://www.opendronemap.org}}, a user-friendly web interface for the widely recognized OpenDroneMap (ODM) toolkit.
It incorporates various well-tested open-source projects such as OpenSfM\footnote{\url{https://opensfm.org}} and OpenMVS\footnote{\url{https://github.com/cdcseacave/openMVS}} for the computation.
By leveraging WebODM, we generate 3D dense reconstructions of the environment, represented through detailed point clouds (\reffig{fig:odm-diyd}) and meshed models.
Two types of recordings are needed for WebODM: 360° videos in equirectangular format and images from a mapping flight. To process the videos, individual frames with maximal information and minimal visual imperfections are extracted. Approximately \numrange{100}{150} individual frames are exported at consistent time intervals. These frames are crucial for creating the 3D model. Mapping flight images can be processed immediately to yield an orthophoto.

\subsubsection{PatchMatch-Stereo-Panorama}
\label{sec:patchmatch}
This approach utilizes the PatchMatch Stereo\cite{bleyer2011patchmatchstereo} algorithm in combination with the well-established  visual SLAM (VSLAM) algorithm, OpenVSLAM\cite{sumikura2019openvslam}\cite{murartal2017orbslam2}.
The camera position is initially tracked by the VSLAM algorithm, and the resulting keyframes are used for the depth map calculation.
By incorporating multiple filters for view selection, depth map cleaning, and point cloud pruning, a sufficient amount of stereo information and local consistency is ensured and duplicates are removed \cite{surmann2022patchmatch}.
Through the optimization of the implementation and GPU acceleration, a real-time computation was achieved.
Common problems in the field of depth estimation are reflective, transparent, and textureless surfaces.
This can be observed in \reffig{fig:pmd-diyd}, where the ceiling and major parts of the car are missing.
The incorporation of dense reconstruction in OpenVSLAM has several benefits.
Firstly, the VSLAM algorithm by itself selects images, with a high information density as key frames and provides pose information for them, optimal for stereo matching \cite{surmann2022patchmatch}.
Secondly, because of this, no further preprocessing is necessary, and thirdly we can export the images, transformations, and point cloud for further use.

\subsubsection{NeRFs}
\label{sec:nerfs}
The speedup of instant-ngp piqued our interest in providing an extended representation of the environment from a sparse set of images to assist rescue forces.
To be able to process the equirectangular video, images have to be extracted and then localized by COLMAP.
These images are then used to train the NeRF.
Such a NeRF offers free movement in space, thus allowing rescue forces an unconstrained selection of viewpoints on the ROI.


Nerfstudio\cite{Tancik_2023} was designed with plug-and-play components for implementing NeRF-based methods, this framework makes it easier for researchers and practitioners to incorporate NeRF into a variety of projects. The modular design of Nerfstudio also enables extensive real-time visualization tools in the browser, provides streamlined pipelines for the import of captured in-the-wild data, and offers tools for exporting to video, point cloud, and mesh formats.

The developers of Nerfstudio released their own NeRF model called Nerfacto which combines components from recent research papers to achieve a balance between speed and quality (\reffig{fig:nerf-diyd}). It is not only designed to be both efficient and high-quality but also remains flexible for future changes or adjustments and is therefore used in this paper.

\section{Results}
\label{sec:results}


\begin{figure*}
    \centering
    \begin{subfigure}[b]{0.329\textwidth}
        \includegraphics[width=\textwidth]{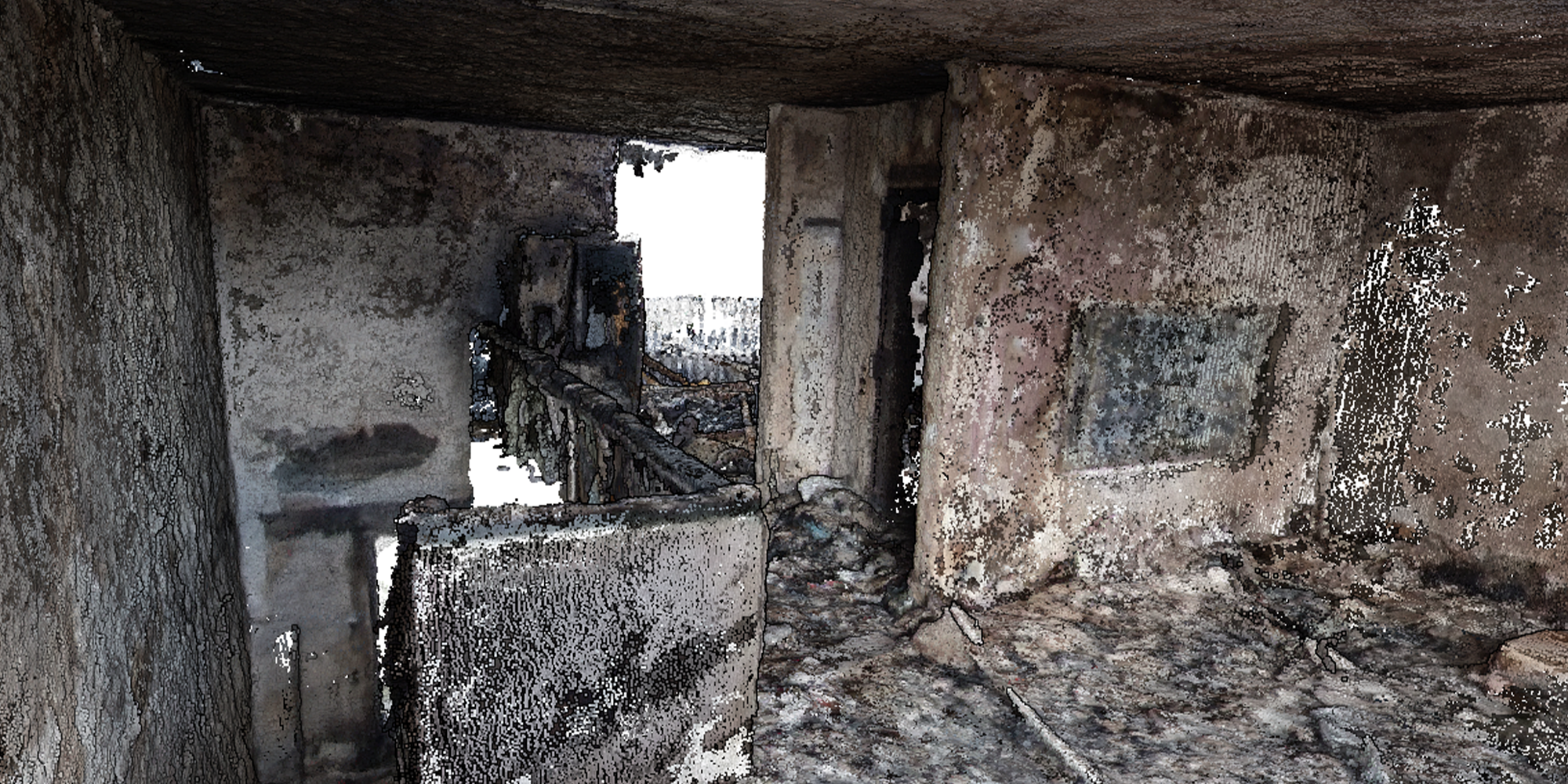}
        \caption{WebODM Point Cloud}
        \label{fig:odm-essen}
    \end{subfigure}
    \hfill
    \begin{subfigure}[b]{0.329\textwidth}
        \includegraphics[width=\textwidth]{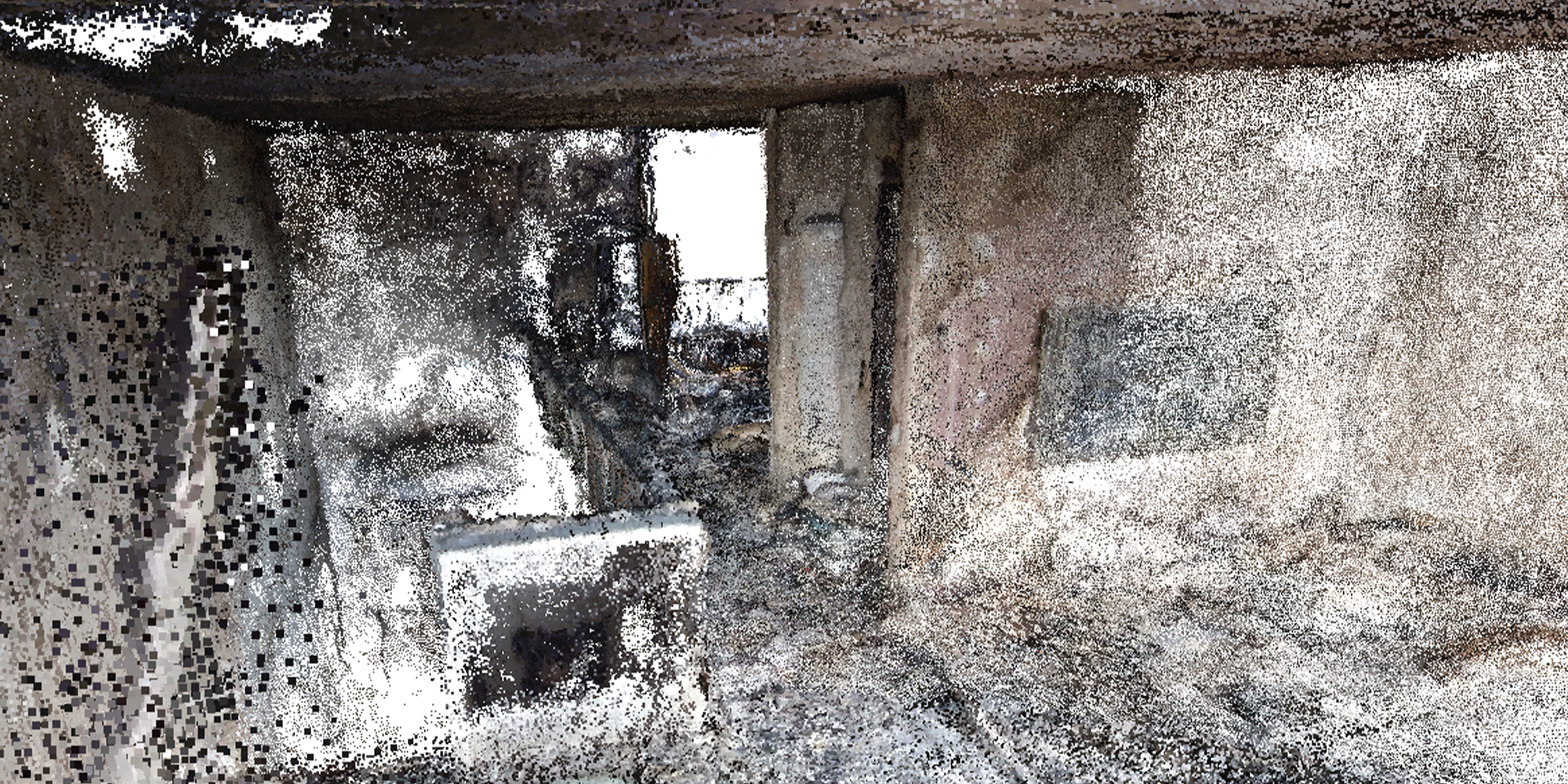}
        \caption{PatchMatch-Stereo-Panorama}
        \label{fig:pmd-essen}
    \end{subfigure}
    \hfill
    \begin{subfigure}[b]{0.329\textwidth}
        \includegraphics[width=\textwidth]{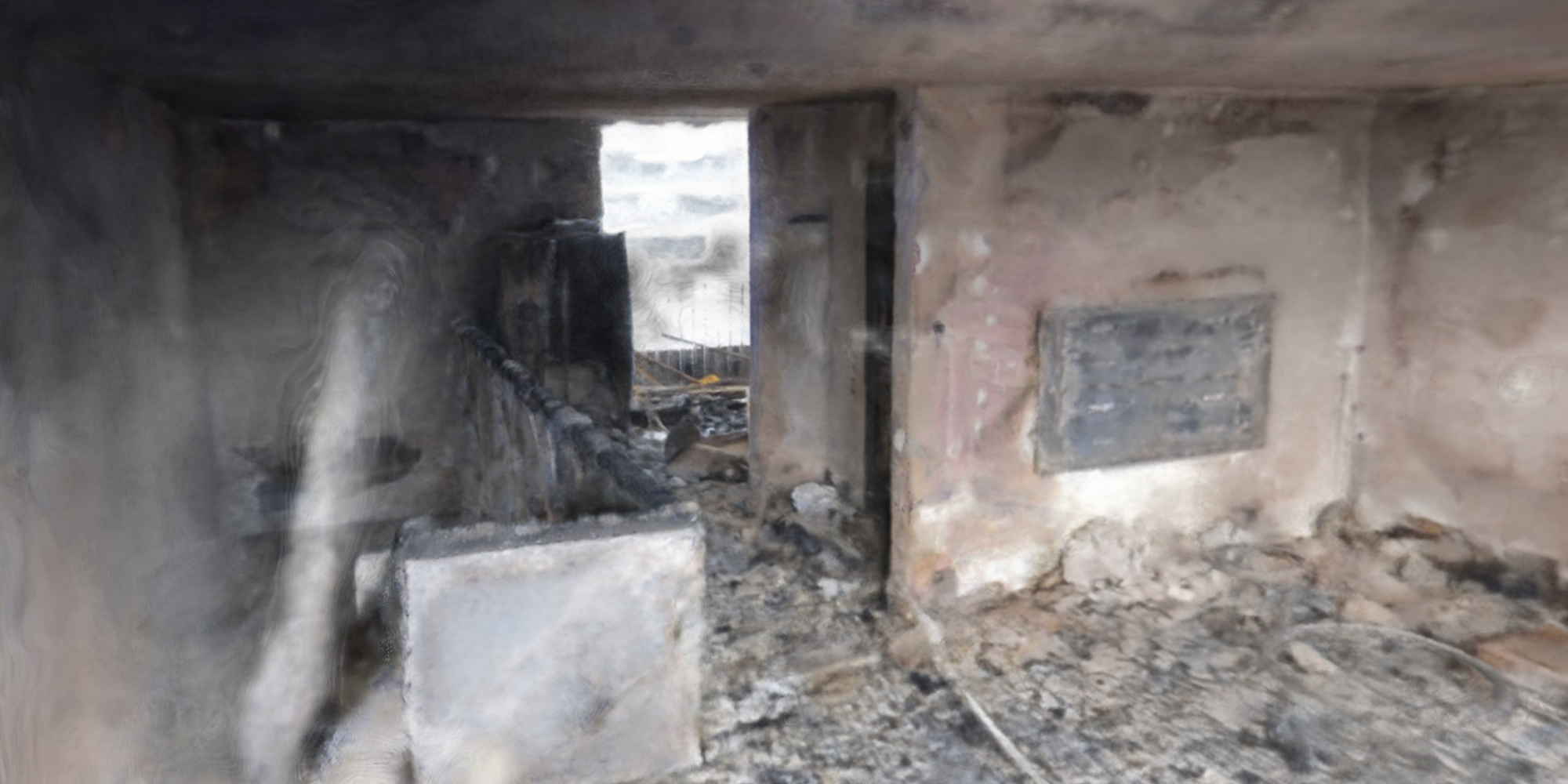}
        \caption{Nerfstudio Nerfacto}
        \label{fig:nerf-essen}
    \end{subfigure}
    \caption{Evaluation of the different methods on a real-world operation inside a burned-out building in Essen, Germany.}
    \label{fig:essen-results}
\end{figure*}

WebODM enables aerial imagery and 3D reconstruction, but computation time is a limiting factor. Calculating a 3D model from \numrange{100}{150} images takes around an hour on our previously described laptop workstation, and this time increases for larger datasets, which renders it less applicable for situations requiring a fast response. Extracting images from the video is time-consuming and not always error-free. Despite these drawbacks, the 3D reconstructions are of high quality, as evident in \reffig{fig:odm-essen}, and georeferenced for overlay and measurements.
This approach has proven to be effective and user-friendly, making it easily deployable for real-world applications.

The PatchMatch Stereo integration in OpenVSLAM eliminates the need for preprocessing. This saves time and speeds up computation, which can be done in real-time. Reconstruction is built continuously and readily available, but this efficiency may affect quality, as seen in \reffig{fig:pmd-essen}.

NeRFs, like WebODM, require additional preprocessing to handle the data. Firstly, images have to be extracted from the 360° video. However, the most time-consuming part is the additional localization using COLMAP. Another limitation is the high computational demand, especially regarding the requirements for RAM and graphics memory. We couldn't run the Nerfacto model on large datasets (above 500 images) on our laptop workstation. This limitation could also be observed on systems with more VRAM (24 GB), limiting the representable scene's size. On the positive side, the NeRF models offer the highest precision in view generation, as depicted in \reffig{fig:nerf-essen}. With sufficient computational power, views are rendered quickly and continuously improve over time. This is especially valuable as, in comparison to only viewing the video, it allows free movement within the space.

\section{Discussion \& Conclusion}
\label{sec:discussion-conclusion}

Undeniably, rescue missions in statically unstable environments, such as those prevalent post-fires and earthquakes, pose tremendous challenges to both humans and robots. We've introduced a system featuring small UAVs, equipped with a 360° camera and software consisting of visual localization methods and NeRF models. These UAVs, capable of swiftly navigating through operational terrains in under 5 minutes, can localize 360° panoramas with the visual localization methods. Subsequently, utilizing the localized 360° images, 3D models of unparalleled quality are generated via NeRF models, markedly outperforming commonly-used multi-view stereo algorithms like PatchMatch Stereo. Remarkably, a 20 x 70 m industrial hall can be externally and internally surveyed in less than 4 minutes with the 3D model being available for situational analysis within an additional 15 minutes. Although our immediate focus has been ensuring practicability for firefighters, future iterations plan to integrate quantitative evaluation methods. To fully harness our approach’s potential, thorough training of emergency personnel is imperative, even with UAVs like the DJI Avatar, which offer substantial flight assistance. Moreover, optimizing the evaluation, presentation, and convenience of image and model result outputs within a situational awareness system via a modern web-based user interface remains a critical endeavor for enhancing end-user intuitiveness and convenience.





\vspace*{1.5ex} 
\noindent\textit{Acknowledgment:}
This work is funded by the Federal Ministry of Education and Research (BMBF) under 
grant, 13N16478 (E-DRZ), cf. \href{https://rettungsrobotik.de}{https://rettungsrobotik.de}. 
We thank our project partners and collaborators.
\vspace*{-0.5ex} 

\bibliography{literatur.bib}
\bibliographystyle{IEEEtran}

\end{document}